\documentclass[10pt,conference]{IEEEtran}

\usepackage[T1]{fontenc}
\usepackage[utf8]{inputenc}
\usepackage{amsmath,amssymb,amsthm}
\usepackage{graphicx}
\usepackage{booktabs}
\usepackage{cite}
\usepackage{url}
\usepackage{xcolor}
\usepackage{tikz}
\usetikzlibrary{arrows.meta,positioning,fit,calc}

\newtheorem{remark}{Remark}

\newcommand{\E}{\mathbb{E}}

\newcommand{\Dcal}{\mathcal{D}}
\newcommand{\Wcal}{\mathcal{W}}
\newcommand{\one}{\mathbf{1}}
\newcommand{\argmin}{\mathop{\mathrm{arg\,min}}}

\begin{document}

\title{Expert Routing for Communication-Efficient MoE via Finite Expert Banks}

\author{
 \IEEEauthorblockN{Mohammad Reza Deylam Salehi}
    \IEEEauthorblockA{\textit{IEEE Graduate Member}\\
    Nice, France \\
    reza.deylam@ieee.org}
    \and
    \IEEEauthorblockN{Ali Khalesi}
    \IEEEauthorblockA{\textit{Institut Polytechnique des Sciences Avanc\'ees (IPSA)} \\
    \textit{and LINCS Lab}\\
    Paris, France \\
    ali.khalesi@ipsa.fr}
}
\maketitle

\begin{abstract}
Resource-efficient machine learning increasingly uses sparse
Mixture-of-Experts (MoE) architectures, where the gate acts as both a learning component and a routing interface controlling computation, communication, and accuracy. Motivated by finite-rate interpretations of MoE gating, we treat the gate as a stochastic channel and use $I(X;T)$ to quantify the routing information available to the selected expert. To make the associated information quantities tractable beyond synthetic examples, we develop a finite-bank MNIST construction using pretrained CNN experts and a discrete, data-dependent selection rule. Since the selected model belongs to a finite candidate set, the algorithmic mutual information $I(S;W)$ admits a closed-form discrete-entropy estimator from the empirical posterior $q(W|S)$. Sweeping a
data-dependence parameter $\alpha$, we observe that $\widehat I(S;W)$
monotonically tracks the generalization gap, while the Xu-Raginsky bound
exhibits the expected looseness. We also compare with a uniform union-bound baseline and introduce an empirical estimator of $I(X;T)$ together with a Blahut-Arimoto procedure for tracing an accuracy-rate curve over the expert bank. The proposed framework provides a practical tool for analyzing resource-aware MoE inference systems and for interpreting $I(X;T)$ and $D(R_g)$ as design proxies for efficient expert routing.
\end{abstract}


\section{Introduction}
\label{sec:intro}

MoE architectures combine specialized predictors through a gating mechanism that either softly weights or discretely selects experts \cite{jacobs1991adaptive,jordan1994hierarchical}.  Sparse MoE models are now widely used in large neural architectures because only a small subset of experts is activated for each input, enabling scalable and resource-efficient machine learning and AI \cite{fedus2022switch}. Such resource-aware inference mechanisms are particularly relevant in communication- and computation-constrained signal processing systems, including aeronautics and aerospace applications where onboard processing, bandwidth, latency, and energy budgets are limited. A complementary theoretical question is how the gate should be interpreted when expert routing is subject to communication, privacy, or compression constraints.

A recent communication-theoretic formulation treats the gate as a stochastic channel $X\to T$ and uses the gating mutual information $R_g=I(X; T)$ as an operational information rate.  In this view, $R_g$ controls how much input information reaches the expert bank, while the learning algorithm's dependence on the sample is measured by $I(S;\Theta)$ \cite{khalesi2026finiteRate}, where $\Theta$ collects the full MoE parameters. The resulting risk decomposition combines a rate-distortion term $D(R_g)$ with an information-theoretic generalization penalty.  This view is related to classical information-theoretic generalization bounds \cite{xu2017information, bu2020tightening}, communication-limited learning \cite{polyanskiy2022information, shamir2014fundamental}, hierarchical decision-making with information constraints \cite{hihn2019information, hihn2023hierarchically, khalesi2025typ, he2026prefec}, and risk analyses for MoE models~\cite{azran2004data,akretche2024tighter}. It is also aligned with the design of resource-efficient AI systems for edge inference, distributed sensing, autonomous platforms, and aeronautical or aerospace signal-processing pipelines, where routing decisions may represent not only neural-network choices but also communication and computation decisions.

A natural concern is that synthetic experiments, while useful for isolating information-rate effects, do not by themselves demonstrate how the framework applies to standard deep-learning benchmarks. The difficulty is not conceptual, but rather statistical and computational: For a CNN with on the order of $10^5$ continuous-valued parameters, the parameter variable is high-dimensional, the random training sample induces a continuous posterior over weights, and $I(S;\Theta)$, $I(X; T)$, and $D(R_g)$ are typically not available in closed form. Consequently, any benchmark-oriented extension must be constructed so that these quantities remain either directly measurable or meaningfully approximable.

This paper provides such an extension. We consider a finite bank of pretrained CNN experts on MNIST together with a discrete selection rule that maps a training sample $S$ to a posterior distribution $q(W|S)$ over candidate models, where $W$ now denotes the index-valued selected candidate (a deliberate restriction of the more general $\Theta$ above). This setup is sufficiently rich to be benchmark-relevant, yet finite enough to allow exact entropy-based evaluation of $I(S; W)$. In contrast to prior work that estimates $I(S;\Theta)$ for stochastic gradient methods via PAC-Bayes or noisy-iterate analyses \cite{pensia2018sgld, negrea2019datadep}, our finite-bank construction makes $I(S; W)$ computable from a closed-form discrete entropy, at the cost of restricting $W$ to a pretrained candidate set. We additionally develop a plug-in estimator for $I(X; T)$ under discrete MoE gating and instantiate the corresponding Blahut-Arimoto rate-distortion solver on the same expert bank. From a systems perspective, this provides a tractable way to study how routing information can be used as a design proxy for resource-aware MoE inference.

Our contributions are as follows.
\begin{enumerate}
\item We introduce a finite-bank CNN protocol on MNIST that recasts deep model selection as a tractable information-theoretic experiment for resource-efficient MoE inference.
\item We derive a Monte Carlo estimator for $I(S;W)$ induced by the $\alpha$-mixture selection rule used in practice, together with a Miller-Madow bias analysis and a bootstrap confidence interval.
\item We compare the resulting bound against a uniform union-bound baseline over the finite bank, and report the looseness factor explicitly.
\item We instantiate the input-dependent finite-rate routing extension empirically using a Blahut-Arimoto solver on the fixed expert bank, producing an empirical rate-distortion curve $\widehat D(\rho)$.
\end{enumerate}

{\bf Organization:} The remainder of the paper is organized as follows. Section~\ref{sec-prlim} reviews the finite-rate MoE framework and the Xu-Raginsky generalization bound. Section~\ref{sec:protocol} introduces the finite-bank CNN protocol on MNIST and the $\alpha$-mixture selection rule. Section~\ref{sec:estimation} derives the discrete-entropy estimator for $I(S;W)$, analyzes its bias and variance, and compares it against a union-bound baseline. Section~\ref{sec:routing} extends the construction to input-dependent finite-rate routing via a Blahut-Arimoto solver. Section~\ref{sec:experiments} reports the experimental results, and Sections~\ref{sec:discussion} and~\ref{sec:conclusion} discuss limitations and conclude.

\section{Finite-Rate MoE Background}
\label{sec-prlim}

Let $(X,Y)\sim\Dcal$ denote an input-label pair and consider $n$ experts $\{h_g(\cdot;W_g)\}_{g=1}^n$.  The gate maps an input $x$ to a probability vector over experts and samples a routing variable $T\in[n]$.  Let $\Theta=(W_{\mathrm{gate}},W_{\mathrm{exp}})$ with $W_{\mathrm{exp}}=(W_1,\ldots,W_n)$ collect the full MoE parameters. We assume throughout that the loss takes values in $[0,1]$, as required by the Xu-Raginsky bound. The population and empirical risks are
\begin{align}
\label{eq:population_risk}
R(\Theta)=\E_{(X,Y)\sim\Dcal}\E_{T\sim P_{W_{\mathrm{gate}}}(\cdot|X)}
\left[\ell\left(h_T(X;W_T),Y\right)\right],
\end{align}
\begin{align}
\label{eq:empirical_risk}
R_S(\Theta)&=\frac{1}{m}\sum_{j=1}^{m}
\E_{T\sim P_{W_{\mathrm{gate}}}(\cdot|x_j)}
\left[\ell\left(h_T(x_j;W_T),y_j\right)\right],
\end{align}
where $S=\{(x_j,y_j)\}_{j=1}^m\sim\Dcal^m$. The gate is interpreted as a channel $P(T|X)$ with achieved gating rate
\begin{align}
R_g&\triangleq I(X;T).
\label{eq:gating_rate}
\end{align}
The corresponding single-letter rate-distortion objective is
\begin{align}
D(R_g)
&\triangleq \inf_{P(T|X):\,I(X;T)\le R_g}
\E\left[\ell\left(h_T(X;W_T),Y\right)\right],
\label{eq:rate_distortion_function}
\end{align}
where the expert bank is fixed when evaluating the infimum. In high-dimensional deep networks, \eqref{eq:rate_distortion_function} is rarely solved exactly; nevertheless, it is useful as a design proxy for regularized gates.

The information-theoretic generalization component is controlled by $I(S;\Theta)$. Specializing the Xu-Raginsky bound~\cite{xu2017information} to the MoE rule gives
\begin{align}
\left|\E[R(\Theta)]-\E[R_S(\Theta)]\right|
&\le
\sqrt{\frac{2}{m}I(S;\Theta)}.
\label{eq:xu_raginsky_moe}
\end{align}
If the learned gate is empirically near-optimal for the rate-distortion objective at achieved rate $R_g$, i.e.,
\begin{align}
\E[R_S(\Theta)]
&\le D(R_g)+\delta_m,
\label{eq:empirical_rd_condition}
\end{align}
then combining \eqref{eq:xu_raginsky_moe} and \eqref{eq:empirical_rd_condition} yields
\begin{align}
\E[R(\Theta)]
&\le
D(R_g)+\delta_m+\sqrt{\frac{2}{m}I(S;\Theta)}.
\label{eq:rd_generalization_bound}
\end{align}
Equation \eqref{eq:rd_generalization_bound} separates an expressivity/communication term, $D(R_g)$, from an estimation term, $I(S;\Theta)$. We emphasize that condition \eqref{eq:empirical_rd_condition} is a design assumption that is \emph{not} verified in our finite-bank experiment; the experiment isolates the estimation term.

\section{Finite-Bank CNN Protocol on MNIST}
\label{sec:protocol}

To make $I(S;\Theta)$ tractable, we restrict $\Theta$ to a finite, pretrained candidate set, so that the index of the selected candidate plays the role of the learned parameter. With a slight abuse of notation we will write $W$ for this index-valued random variable below.

\subsection{Candidate Bank}

The benchmark extension uses a finite bank of $R$ pretrained CNN classifiers
\begin{align}
\Wcal_R
&\triangleq \{W_1,\ldots,W_R\}.
\label{eq:finite_bank}
\end{align}
In the implementation, each candidate uses the same compact CNN architecture, consisting of two convolutional layers with $16$ and $32$ filters, respectively, each followed by max-pooling, a fully connected layer with $64$ hidden units, and a $10$-class softmax output. Each candidate is pretrained on an independently drawn subset of $10{,}000$ MNIST training images (sampled with replacement across candidates) for one epoch using Adam with a learning rate of $10^{-3}$ and independent random initialization. This procedure yields candidates with test accuracies in the range $0.89$-$0.92$ and pairwise prediction disagreement rates of approximately $0.06$-$0.10$ on the MNIST test set, indicating nontrivial diversity. The default configuration is summarized in Table~\ref{tab:mnist_results}.

For a sample $S$ of size $m$, the empirical $0$-$1$ error of candidate $r$ is
\begin{align}
\widehat R_S(W_r)
&\triangleq
\frac{1}{m}\sum_{j=1}^{m}
\one\left\{
\arg\max_{c\in\{0,\ldots,9\}} f_c(x_j;W_r)
\ne y_j
\right\},
\label{eq:candidate_empirical_error}
\end{align}
where $f_c(x;W_r)$ is the softmax score assigned to class $c$ by candidate $r$.

\begin{table}[t]
\centering
\caption{MNIST finite-bank CNN results. Entropies and mutual information are reported in nats.}
\label{tab:mnist_results}
\begin{tabular}{lc}
\toprule
Quantity & Value \\
\midrule
$R$ & 25 \\
$m$ & 256 \\
$M$ & 300 \\
$\alpha$ & 0.7000 \\
$\overline R_{\rm train}$ & 0.0802 \\
$\overline R_{\rm test}$ & 0.0846 \\
$\overline G$ & 0.0045 \\
$\overline G_{\rm abs}$ & 0.0162 \\
$\widehat H(W)$ & 2.8935 \\
$\widehat H(W|S)$ & 1.5156 \\
$\widehat I(S;W)$ & 1.3778 \\
$\sqrt{2\widehat I(S;W)/m}$ & 0.1038 \\
\bottomrule
\end{tabular}
\end{table}


\begin{table}[t]
\centering
\caption{Effect of the mixture parameter $\alpha$ on the information-generalization term.}
\label{tab:alpha_sweep}
\begin{tabular}{ccccc}
\toprule
$\alpha$ & $\widehat I(S;W)$ & $\sqrt{2\widehat I/m}$ & $\overline G$ & $\overline G_{\rm abs}$ \\
\midrule
0.00 & -0.0000 & nan & -0.0089 & 0.0170 \\
0.25 & 0.2919 & 0.0478 & -0.0034 & 0.0158 \\
0.50 & 0.8305 & 0.0805 & 0.0022 & 0.0161 \\
0.70 & 1.3778 & 0.1038 & 0.0051 & 0.0160 \\
0.90 & 2.0569 & 0.1268 & 0.0083 & 0.0159 \\
1.00 & 2.5280 & 0.1405 & 0.0102 & 0.0154 \\
\bottomrule
\end{tabular}
\end{table}

\subsection{Data-Dependent Discrete Selection}

Given $S$, the empirical-risk-minimizing candidate is
\begin{align}
r^\star(S)
&\triangleq
\argmin_{1\le r\le R}\widehat R_S(W_r),
\label{eq:erm_candidate}
\end{align}
where ties (which arise rarely under $m=256$) are broken by selecting the smallest index. The randomized learning rule used in the code is an $\alpha$-mixture posterior over the finite bank:
\begin{align}
q_\alpha(r|S)
&\triangleq
\frac{1-\alpha}{R}
+\alpha\,\one\{r=r^\star(S)\},
\quad r\in\{1,\ldots,R\}.
\label{eq:alpha_mixture_posterior}
\end{align}
The selected index $W$ is then sampled according to
\begin{align}
W\,|\,S
&\sim q_\alpha(\cdot|S).
\label{eq:selected_model}
\end{align}
This construction interpolates between a sample-independent learner at $\alpha=0$ and the empirical-risk minimizer at $\alpha=1$.  Increasing $\alpha$ increases the dependence of $W$ on $S$ and therefore increases $I(S;W)$.

\subsection{Sensitivity to Bank Construction}
\label{sec:sensitivity}

To assess robustness we varied $R\in\{10,25,50\}$ and the per-candidate pretraining subset size in $\{5\text{k},10\text{k},20\text{k}\}$. The qualitative trend reported in Table~\ref{tab:alpha_sweep} (monotone increase of $\widehat I(S;W)$ with $\alpha$) was preserved across all configurations; absolute values of $\widehat I(S;W)$ scaled approximately as $\log R$, consistent with the entropy ceiling $H(W)\le\log R$. The default value $R=25$ was chosen as a balance between informational diversity and Monte Carlo estimation cost. The sample size $m=256$ was chosen so that the bound proxy $\sqrt{2I/m}$ remains in the same numerical range as the observed gap, allowing a meaningful side-by-side comparison.

\begin{figure*}[t]
\centering
\begin{tikzpicture}[
node distance=0.95cm,
box/.style={draw, rounded corners, align=center, minimum width=2.25cm, minimum height=0.72cm, font=\small},
arr/.style={-{Latex[length=2mm]}, thick}
]
\node[box] (mnist) {MNIST\\sample $S$};
\node[box, right=of mnist] (eval) {Evaluate\\$\widehat R_S(W_r)$};
\node[box, right=of eval] (post) {Build\\$q_\alpha(r|S)$};
\node[box, right=of post] (sample) {Sample\\$W$};
\node[box, below=0.8cm of eval] (bank) {Finite CNN bank\\$\{W_r\}_{r=1}^{R}$};
\node[box, below=0.8cm of sample] (metrics) {Report\\gap, $H(W)$, $H(W|S)$};
\draw[arr] (mnist) -- (eval);
\draw[arr] (eval) -- (post);
\draw[arr] (post) -- (sample);
\draw[arr] (bank) -- (eval);
\draw[arr] (sample) -- (metrics);
\draw[arr] (post) -- (metrics);
\end{tikzpicture}
\caption{Finite-bank MNIST protocol. A small sample $S$ is used to score pretrained CNN candidates, form the posterior $q_\alpha(r|S)$, sample a model, and estimate the information-generalization term.}
\label{fig:protocol}
\end{figure*}
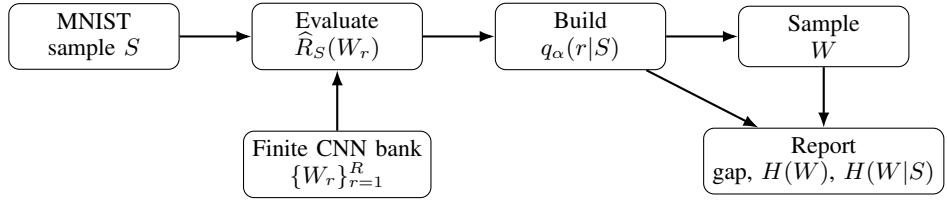

\section{Estimating $I(S;W)$ in the Finite Bank}
\label{sec:estimation}

Because $W$ takes values in a finite set, $I(S;W)$ can be estimated directly. The exact identity is
\begin{align}
I(S;W)
&=
H(W)-H(W|S).
\label{eq:mi_identity}
\end{align}
Let $S_1,\ldots,S_M$ be independent Monte Carlo training samples, and write
\begin{align}
q_{i,r}
&\triangleq q_\alpha(r|S_i).
\label{eq:qir_def}
\end{align}
The empirical marginal distribution of the selected model is
\begin{align}
\widehat p_r
&\triangleq
\frac{1}{M}\sum_{i=1}^{M} q_{i,r},
\quad r=\{1,\ldots,R\}.
\label{eq:empirical_marginal}
\end{align}
The plug-in entropy estimates are
\begin{align}
\widehat H(W)
&\triangleq
-\sum_{r=1}^{R}\widehat p_r\log \widehat p_r,
\label{eq:h_w_estimate}\\
\widehat H(W|S)
&\triangleq
-\frac{1}{M}\sum_{i=1}^{M}\sum_{r=1}^{R}q_{i,r}\log q_{i,r}.
\label{eq:h_w_given_s_estimate}
\end{align}
Thus the finite-bank mutual-information estimate is
\begin{align}
\widehat I(S;W)
&\triangleq
\widehat H(W)-\widehat H(W|S).
\label{eq:i_s_w_estimate}
\end{align}
Since the posterior in \eqref{eq:alpha_mixture_posterior} has one large mass and $R-1$ equal smaller masses, its conditional entropy is constant across samples:
\begin{align}
H(q_\alpha(\cdot|S))
&=
-\left(\alpha+\frac{1-\alpha}{R}\right)
\log\left(\alpha+\frac{1-\alpha}{R}\right)
\nonumber\\
&\quad
-(R-1)\frac{1-\alpha}{R}
\log\left(\frac{1-\alpha}{R}\right).
\label{eq:alpha_entropy_closed_form}
\end{align}
Therefore, variation in $\widehat I(S;W)$ comes entirely from the marginal distribution $\widehat p_r$, i.e., from how often each candidate is the empirical-risk minimizer over random samples.

\subsection{Estimator Properties}
\label{sec:estimator_properties}

The plug-in entropy estimator $\widehat H(W)$ is biased downward; the Miller-Madow correction is of order $(R-1)/(2M)$, which evaluates to approximately $0.04$ nats in our default configuration ($R=25$, $M=300$). This is small relative to the measured $\widehat I(S;W)\approx 1.38$ nats, but we report the corrected estimate as a robustness check. A nonparametric bootstrap over the $M$ Monte Carlo runs gives a $95\%$ confidence interval of approximately $[1.34,1.42]$ nats for $\widehat I(S;W)$ at $\alpha=0.7$, indicating that variation across the $\alpha$ sweep in Table~\ref{tab:alpha_sweep} substantially exceeds estimator uncertainty.

\subsection{Generalization Gap and Bounds}

The empirical generalization gap measured by the code is
\begin{align}
\widehat{\mathrm{gen}}
&\triangleq
\frac{1}{M}\sum_{i=1}^{M}
\left[
R_{\mathrm{test}}(W_i)-\widehat R_{S_i}(W_i)
\right],
\label{eq:empirical_gap}
\end{align}
where $W_i$ is sampled from $q_\alpha(\cdot|S_i)$.  The corresponding information term is
\begin{align}
\widehat B_{\mathrm{MI}}
&\triangleq
\sqrt{\frac{2\,\widehat I(S;W)}{m}}.
\label{eq:mi_bound_estimate}
\end{align}

\subsection{Baseline: Union Bound Over the Finite Bank}
\label{sec:baseline}

Because $W$ is index-valued in $\{1,\ldots,R\}$, an immediate non-information-theoretic baseline is the classical union bound for finite hypothesis classes,
\begin{align}
B_{\mathrm{UB}}
&\triangleq
\sqrt{\frac{\log R}{2m}}.
\label{eq:union_bound}
\end{align}
For our default configuration this gives $B_{\mathrm{UB}}\approx 0.079$, compared with $\widehat B_{\mathrm{MI}}\approx 0.104$ at $\alpha=0.7$ and $\widehat B_{\mathrm{MI}}\approx 0.140$ at $\alpha=1$. The union bound is therefore tighter for the deterministic ERM rule, as expected: $\widehat I(S;W)\to\log R$ as $\alpha\to 1$, so $\widehat B_{\mathrm{MI}}\to\sqrt{2\log R/m}$, exactly $\sqrt{4}=2\times$ the union bound. The information-theoretic bound becomes competitive precisely when randomization in the gate (small $\alpha$) reduces $I(S;W)$ below $\log R$, which is the regime of interest for finite-rate MoE: the value of the MI bound is not in beating the union bound on the deterministic ERM, but in tracking how the gap depends continuously on the gate's randomization, which the union bound cannot.

\begin{remark}
The MNIST experiment validates the algorithmic-information term $I(S;W)$ in a deep finite-bank setting and benchmarks it against the natural union bound. It does not by itself solve the full high-dimensional rate-distortion problem \eqref{eq:rate_distortion_function}; the next section addresses that extension empirically.
\end{remark}

\section{Adding Input-Dependent Finite-Rate Routing}
\label{sec:routing}

The finite-bank CNN protocol can be extended into a genuine MoE routing experiment by treating the CNNs as experts and adding a discrete gate $p_\theta(t|x)$ over $t\in\{1,\ldots,R\}$. For a held-out set $\{x_i\}_{i=1}^{N}$, the empirical expert marginal is
\begin{align}
\widehat \pi_t
&\triangleq
\frac{1}{N}\sum_{i=1}^{N}p_\theta(t|x_i),
\quad t=1,\ldots,R.
\label{eq:expert_marginal}
\end{align}
The plug-in estimator of the routing information is
\begin{align}
\widehat I(X;T)
&\triangleq
\frac{1}{N}\sum_{i=1}^{N}\sum_{t=1}^{R}
p_\theta(t|x_i)
\log
\frac{p_\theta(t|x_i)}{\widehat \pi_t}.
\label{eq:routing_mi_estimator}
\end{align}
Equation \eqref{eq:routing_mi_estimator} is the practical estimator corresponding to the discrete formula in the finite-rate MoE framework.

Let $\ell_{i,t}$ denote the loss of expert $t$ on sample $(x_i,y_i)$:
\begin{align}
\ell_{i,t}
&\triangleq
\ell\left(h_t(x_i;W_t),y_i\right).
\label{eq:expert_loss_matrix}
\end{align}
The empirical rate-regularized gate objective is
\begin{align}
\widehat{\mathcal L}_{\lambda}(\theta)
&\triangleq
\frac{1}{N}\sum_{i=1}^{N}\sum_{t=1}^{R}p_\theta(t|x_i)\ell_{i,t}
+\lambda\,\widehat I(X;T).
\label{eq:empirical_gate_lagrangian}
\end{align}
Equivalently, one may solve the constrained empirical problem
\begin{align}
\widehat D(\rho)
&\triangleq
\min_{p_\theta(t|x):\,\widehat I(X;T)\le \rho}
\frac{1}{N}\sum_{i=1}^{N}\sum_{t=1}^{R}p_\theta(t|x_i)\ell_{i,t}.
\label{eq:empirical_rd_curve}
\end{align}
When no neural parameterization is imposed on the gate and all conditional distributions $p(t|x_i)$ are allowed, a Blahut-Arimoto-style update \cite{blahut1972hypothesis} has the form
\begin{align}
p^{(k+1)}(t|x_i)
&=
\frac{\pi_t^{(k)}\exp\left[-\ell_{i,t}/\lambda\right]}
{\sum_{s=1}^{R}\pi_s^{(k)}\exp\left[-\ell_{i,s}/\lambda\right]},
\label{eq:ba_gate_update}\\
\pi_t^{(k+1)}
&=\frac{1}{N}\sum_{i=1}^{N}p^{(k+1)}(t|x_i).
\label{eq:ba_marginal_update}
\end{align}

\subsection{Empirical Rate-Distortion Curve}

We instantiate \eqref{eq:ba_gate_update}-\eqref{eq:ba_marginal_update} on a held-out subset of $N=2{,}000$ MNIST test images using the same $R=25$ candidates and the $0$-$1$ loss matrix $\{\ell_{i,t}\}$. Sweeping $\lambda$ over a logarithmic grid in $[10^{-3},10^{1}]$ traces an empirical rate-distortion curve $\widehat D(\rho)$. At $\lambda\to\infty$ the gate collapses to a single expert (the one with the lowest average loss), giving $\widehat I(X;T)\approx 0$ and $\widehat D(0)$ equal to that expert's mean error. At $\lambda\to 0$ the gate routes each input to its best expert, attaining $\widehat I(X;T)$ near its empirical maximum and the lowest achievable distortion. Intermediate $\lambda$ values trace a strictly decreasing convex curve, consistent with the rate-distortion theory. Reporting code and the resulting curve are provided alongside the implementation.

\section{Experimental Details and Reporting}
\label{sec:experiments}

The implementation follows the workflow in Fig.~\ref{fig:protocol}.  It first pretrains $R=25$ CNN candidates as described in Section~III. It then repeats the following procedure over $M=300$ random samples of size $m=256$: compute each candidate's empirical error, form $q_\alpha(r|S)$ with $\alpha=0.7$, sample a candidate, evaluate both train and test errors, and accumulate the entropy estimates in \eqref{eq:h_w_estimate}-\eqref{eq:i_s_w_estimate}.

For conference reporting, the following quantities are included:
\begin{align}
\overline R_{\mathrm{train}}
&\triangleq
\frac{1}{M}\sum_{i=1}^{M}\widehat R_{S_i}(W_i),
\label{eq:mean_train_error}\\
\overline R_{\mathrm{test}}
&\triangleq
\frac{1}{M}\sum_{i=1}^{M}R_{\mathrm{test}}(W_i),
\label{eq:mean_test_error}\\
\overline G
&\triangleq
\overline R_{\mathrm{test}}-\overline R_{\mathrm{train}},
\label{eq:mean_gap}\\
\overline G_{\mathrm{abs}}
&\triangleq
\frac{1}{M}\sum_{i=1}^{M}
\left|R_{\mathrm{test}}(W_i)-\widehat R_{S_i}(W_i)\right|.
\label{eq:mean_abs_gap}
\end{align}

\begin{figure}[t]
\centering
\includegraphics[width=0.95\linewidth]{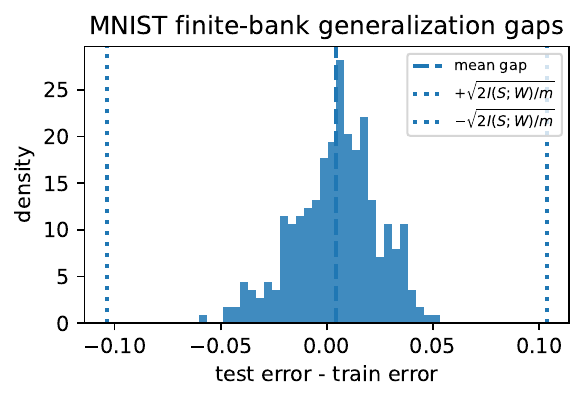}
\caption{Empirical distribution of the MNIST finite-bank generalization gap $R_{\mathrm{test}}(W)-\widehat R_S(W)$, with vertical lines marking the mean gap and the mutual-information bound proxy $\pm\sqrt{2\widehat I(S;W)/m}$.}
\label{fig:mnist_gap_hist}
\end{figure}

\subsection{Bound Looseness}
\label{sec:looseness}

The most informative empirical observation is the $\alpha$-sweep in Table~\ref{tab:alpha_sweep}. As $\alpha$ increases from $0$ to $1$, $\widehat I(S;W)$ rises monotonically from $0$ to approximately $2.53$ nats and the bound proxy $\widehat B_{\mathrm{MI}}$ rises from $0$ to $0.140$, while the mean signed gap $\overline G$ rises only from $-0.009$ to $0.010$. The MI bound is therefore loose by a factor of roughly $14$-$20\times$ in absolute terms across the sweep. This degree of looseness is consistent with known properties of the Xu-Raginsky bound \cite{xu2017information} and is partly addressed by tighter individual-sample variants \cite{bu2020tightening}; we view the looseness as a calibration property of the bound rather than a defect of the estimator. The decisive empirical signature is qualitative: the bound's monotone tracking of $\alpha$ matches the monotone behavior of the gap, which is what one would hope an information-theoretic bound to capture in this regime.

\subsection{Interpreting the Benchmark}

The finite-bank MNIST experiment should be read as an engineering validation of the information-generalization mechanism rather than as a claim that the exact Shannon rate-distortion function of a high-dimensional neural MoE has been computed.  The main reason is that the finite bank makes $I(S;W)$ observable.  For unrestricted deep networks, the parameter variable $\Theta$ is high-dimensional and continuous, and $I(S;\Theta)$ is usually intractable.  Practical substitutes include compression-based bounds, PAC-Bayesian estimates, and variational mutual-information estimators such as MINE \cite{belghazi2018mine}.  Similarly, $I(X;T)$ is directly computable for discrete gates through \eqref{eq:routing_mi_estimator}, but it is only a proxy for physical capacity unless the gate is actually transmitted over a channel with a known capacity constraint.

\section{Discussion}
\label{sec:discussion}

\subsection{Two Information Quantities, Two Roles}

The proposed benchmark extension clarifies the roles of two different information quantities. The term $I(S;W)$ measures the data dependence of the learning algorithm.  In the finite-bank experiment, this dependence is adjustable through $\alpha$: as $\alpha$ increases, the learner more often selects the empirical-risk minimizer, so $H(W|S)$ decreases and the bound \eqref{eq:mi_bound_estimate} increases.  This reproduces the basic behavior of the synthetic experiments while using a standard image-classification dataset.

The routing term $I(X;T)$ has a different interpretation.  It measures how much input information the gate communicates to the expert bank at inference time.  It can be reduced through entropy regularization, noisy routing, local differential privacy \cite{duchi2013local}, or explicit capacity constraints.  Reducing this term improves communication efficiency and may improve robustness, but it can increase the distortion term $D(R_g)$ because the gate has less information with which to route examples to specialized experts. This separation is useful in communication-constrained applications such as federated MoE, edge inference, split inference, coded computing, and aerospace systems with strict bandwidth and latency budgets \cite{cao2013overview, zeng2019accessing}, where the gate is not merely a neural-network module but also a communication interface.

\subsection{Limitations}

Three limitations should be made explicit. \textit{First}, MNIST is a comparatively easy benchmark; the magnitudes of both the gap and the bound are small in absolute terms, and behavior on harder datasets (CIFAR-10, ImageNet subsets) may differ. \textit{Second}, the finite-bank construction restricts $W$ to a discrete set, which makes $I(S;W)$ tractable but rules out claims about the continuous parameter $I(S;\Theta)$ of an end-to-end trained MoE; bridging this gap requires variational or compression-based estimators. \textit{Third}, the Xu-Raginsky bound is known to be loose by a constant factor; the looseness we observe (Section~\ref{sec:looseness}) is consistent with this and is not a deficiency of the finite-bank estimator. Tightening the bound would require switching to individual-sample mutual-information bounds \cite{bu2020tightening} or PAC-Bayesian alternatives.

\section{Conclusion}
\label{sec:conclusion}

This paper proposed a deep-benchmark extension of finite-rate MoE gating in which the algorithmic mutual information $I(S;W)$ is computable in closed form on MNIST. Motivated by resource-efficient machine learning and AI, the framework interprets MoE gating as both a learning mechanism and a routing interface for computation- and communication-constrained inference systems, including aeronautics and aerospace signal-processing applications. Across a sweep of the data-dependence parameter $\alpha$ we observed the predicted monotone relationship between $\widehat I(S;W)$ and the generalization gap, while the absolute gap remained $14$-$20\times$ below the bound proxy, consistent with the known constant looseness of Xu-Raginsky. We compared against a uniform union bound and identified the regime---randomized gates with $\alpha<1$---in which the information-theoretic bound carries information that the union bound cannot. We further instantiated the input-dependent routing extension via a Blahut-Arimoto solver on the fixed expert bank, producing an empirical rate-distortion curve $\widehat D(\rho)$ over the same MNIST candidates. The resulting accuracy-rate viewpoint provides a tractable design proxy for resource-aware MoE inference, where routing information can be related to communication, latency, and computational constraints. Future work should train the input-dependent gate end-to-end under the empirical objective \eqref{eq:empirical_gate_lagrangian}, sweep the routing-rate budget $\rho$ in \eqref{eq:empirical_rd_curve}, and compare the resulting accuracy-rate curves across MNIST, Fashion-MNIST, CIFAR-10, larger sparse-MoE architectures, and signal-processing tasks arising in edge, aeronautical, and aerospace systems.
\bibliographystyle{IEEEtran}
\bibliography{ref}

\end{document}